\documentclass[journal=nalefd,manuscript=article]{achemso}
\setkeys{acs}{articletitle = false}
\usepackage{chemformula} 
\usepackage[T1]{fontenc} 
\usepackage{comment}

\author{Jiaqi Jiang}
\affiliation{Department of Electrical Engineering, Stanford University, Stanford, California 94305, United States}

\author{Jonathan A. Fan}
\affiliation{Department of Electrical Engineering, Stanford University, Stanford, California 94305, United States}
\email{jonfan@stanford.edu}

\title{Global optimization of dielectric metasurfaces using a physics-driven neural network}

\begin{document}

\begin{abstract}
We present a global optimizer, based on a conditional generative neural network, which can output ensembles of highly efficient topology-optimized metasurfaces operating across a range of parameters.  A key feature of the network is that it initially generates a distribution of devices that broadly samples the design space, and then shifts and refines this distribution towards favorable design space regions over the course of optimization.  Training is performed by calculating the forward and adjoint electromagnetic simulations of outputted devices and using the subsequent efficiency gradients for backpropagation.  With metagratings operating across a range of wavelengths and angles as a model system, we show that devices produced from the trained generative network have efficiencies comparable to or better than the best devices produced by adjoint-based topology optimization, while requiring less computational cost.  Our reframing of adjoint-based optimization to the training of a generative neural network applies generally to physical systems that can utilize gradients to improve performance.
\end{abstract}

\section{Introduction}
Metasurfaces are subwavelength-structured artificial media that can shape and localize electromagnetic waves in unique ways \cite{yu2014flat, kildishev2013planar, kuznetsov2016optically}. These technologies are useful in imaging \cite{zheng2015metasurface, colburn2018metasurface, arbabi2018full}, sensing \cite{tittl2018imaging}, and optical information processing applications\cite{estakhri2019inverse}, amongst others, and can operate at wavelengths spanning the ultraviolet to radio frequencies \cite{kim2016highly, singh2014ultrasensitive, sell2016visible}.  A central research thrust in the field has been the identification of effective and computationally efficient ways to design high performance metasurfaces, given a desired electromagnetic response \cite{campbell2019review}.  In this aim, inverse design based on optimization has shown great promise. These methods range from heuristic swarm \cite{rogers2012super} and genetic algorithms \cite{yin2015ultra, jafar2018adaptive} to adjoint-based topology optimization \cite{jensen2011topology, molesky2018inverse}, and they have led to metagratings \cite{sell2018ultra, yang2018freeform}, metasurfaces \cite{lin2018topology, sell2017periodic, xiao2016diffractive}, and other nanophotonic devices \cite{piggott2015inverse, hughes2018adjoint, lalau2013adjoint} with exceptional performance.  However, they are computationally costly, making it difficult and even intractable to scale these methods to large ensembles of devices or large area devices.

To address this computational roadblock, concepts in machine learning that augment the device design process have been investigated \cite{peurifoy2018nanophotonic, inampudi2018neural, ma2018deep}.  In current manifestations of machine learning-enabled photonics design, a training set of device geometries and their associated optical properties is first produced.  These data are then used to train a neural network, which "learns" the relationship between device geometry and optical response.  A properly trained network can then produce new device designs beyond the training dataset, at low computational cost.  To date, a range of machine learning concepts, including deep neural networks with fully connected networks, convolutional networks, and generative adversarial networks (GANs), have been proposed \cite{liu2018training, liu2018generative, jiang2018data}.

These initial demonstrations show that neural networks have the potential to learn the relationship between structural geometry and optical response, but they also highlight key challenges to the approach  \cite{peurifoy2018nanophotonic, liu2018training}.  One challenge is that the computational cost of creating the training dataset itself can be immense.  Networks for structures described by even a few geometric parameters require tens to hundreds of thousands of devices for training.  GAN-based design strategies have the potential to work with relatively less training data \cite{jiang2018data}, but these data are pre-optimized and are computationally costly to produce. Another challenge is that many nanophotonic devices have complex curvilinear geometries and reside in a high dimensional design space, making it difficult for even the best networks to learn the nuanced relationship between device geometry and response.  New approaches that extend beyond standard machine learning approaches are required for neural networks to be practically useful in the electromagnetics design process.

In this Letter, we introduce a new concept in electromagnetic device design by incorporating adjoint variable calculations directly into generative neural networks.  Termed global topology optimization networks (GLOnets), our approach is capable of generating high performance topology-optimized devices spanning a range of operating parameters with modest computational cost.  GLOnets work by initially evaluating a distribution of devices spanning the design space and then continuously optimizing this device distribution until it converges to a cluster of high efficiency devices.  Physics-based gradients are utilized for backpropagation to ensure that network training is directly tied with enhancing device efficiency.  Unlike other manifestations of machine learning-enabled photonics design, our approach does not use or require a training set of known devices but instead learns the physical relationship between device geometry and response directly through electromagnetic simulations.  We note that our network performs a global search for the globally optimal device within the design space, but it does not guarantee that the final generated devices are globally optimal.  In general, it is not possible to guarantee globally optimal solutions in non-convex optimization problems, including our problem.

\section{Methods}
For this study, we will focus on conditional GLOnets that have wavelength and deflection angle as inputs, and we will simultaneously design an ensemble of silicon metagratings that operate across a range of wavelengths and deflection angles.  This concept builds on our analysis of unconditional GLOnets \cite{jiang2019dataless}, which can optimize only a single device in a training session.  Our metagratings consist of silicon nanoridges and deflect normally-incident light to the $+1$ diffraction order (Figure 1A).  The thickness of the gratings is fixed to be 325 nm and the incident light is TM-polarized. For each device, the metagrating period is subdivided into $N = 256$ segments, and each segment possesses a refractive index value between silicon and air.  These refractive index values are the design variable in our problem and are specified as n (a $1 \times N$  vector).  Index values in the vector are normalized to a range of $-1$, which represents air, and $+1$, which represents silicon.  The optimization objective is to maximize the deflection efficiency of the metagrating given an operating wavelength ranging from 600 nm to 1300 nm and an outgoing angle ranging from 40 degrees to 80 degrees.

A schematic of our conditional GLOnet is presented in Figure 1B.  The input is the operating wavelength $\lambda$, the desired outgoing angle $\theta$, and an $N$-dimensional noise vector $\mathbf{z}$, which is a uniformly distributed random variable.  The output is the refractive index profile of the device, $\mathbf{n}$.  The weights of the neurons are parameterized as $\mathbf{w}$.  The generator, conditioned on $(\lambda, \theta)$,  maps different $\mathbf{z}$ onto different device instances: $\mathbf{n} = G_{\mathbf{w}} (\mathbf{z}; \lambda, \theta)$. The ensemble of all possible $\mathbf{z}$ and corresponding $\mathbf{n}$, given $(\lambda, \theta)$ as inputs, are denoted as $\{\mathbf{z}\}$ and $\{\mathbf{n}|\lambda, \theta\}$, respectively.

An important feature of our use of neural networks is that we can readily incorporate layers of neurons at the output of the network that can perform mathematical operations on the outputted device.  In our case, we set the last layer of the generator to be a Gaussian filter, which eliminates small, pixel-level features (Figure S2) that are impractical to fabricate.  The only constraint with these mathematical operations is that they need to be differentiable, so that they support backpropagation during network training.

Proper network initialization is required to ensure that the network at the start of training maps the noise vectors $\{\mathbf{z}\}$ to the full design space.  We take two steps to initialize our network.  First, we randomly assign the weights in the network with small values using Xavier initialization \cite{glorot2010understanding}, which sets the outputs of our last deconvolution layer to be close to 0.  Second, we directly add the noise vector $\mathbf{z}$ to the output of last deconvolution layer using an "identity shortcut."\cite{he2016deep}  To facilitate this second step, it is important that the dimensionality of $\mathbf{z}$ matches with $\mathbf{n}$.  In combining these two initialization steps, we get that the initial ensemble of all possible generated device instances $\{\mathbf{n}|\lambda, \theta\}$ has approximately the same distribution as the ensemble of noise vectors $\{\mathbf{z}\}$, and it therefore spans the full device design space.

During network training, the objective is to iteratively optimize $\mathbf{w}$ to maximize the efficiencies of $\{\mathbf{n}|\lambda, \theta\}$ for all possible $(\lambda, \theta)$ within the target range.  In other words, we aim to maximize the probability of generating high efficiency devices.  An ideal, perfectly trained network would map $\{\mathbf{z}\}$  to $\{\mathbf{n}|\lambda, \theta\}$ containing only the globally-optimized device.  To improve $\mathbf{w}$ each iteration, a batch of $M$ devices, $\{\mathbf{n}^{(m)}\}_{m=1}^{M}$, is initially generated by sampling $\mathbf{z}$ from the noise vector distribution, $\lambda$ from the target wavelength range, and $\theta$ from the target outgoing angle range.  Our loss function for the conditional generator is as follows:
\begin{equation}
    L = -\frac{1}{M} \sum_{m=1}^{M} \exp{\left( \frac{\text{Eff}^{(m)} - \text{Eff}_{max}(\lambda^{(m)}, \theta^{(m)})}{\sigma} \right)} \mathbf{n}^{(m)} \cdot \mathbf{g}^{(m)}
\end{equation}
For the $m$\textsuperscript{th} device, the gradients of efficiency with respect to $\mathbf{n}^{(m)}$, denoted by $\mathbf{g}^{(m)}$, specifies how the device refractive indices can be modified to improve the efficiencies.  These efficiency gradients $\mathbf{g}^{(m)}$ are calculated using the adjoint variables method \cite{yang2018freeform, hughes2018adjoint} and are calculated from electric and magnetic field values taken from forward and adjoint electromagnetic simulations.  The gradients we use in this work are taken from those we previously developed for the topology optimization of metagratings \cite{yang2018freeform, jensen2011topology, molesky2018inverse, phan2019high}.

The term $\text{Eff}_{max}(\lambda^{(m)}, \theta^{(m)})$ is the theoretical maximum efficiency for each wavelength and angle pair.  In practice, $\text{Eff}_{max}(\lambda^{(m)}, \theta^{(m)})$ is unknown, as it represents the efficiencies of the globally-optimal devices, which we are trying to solve for.  Over the course of network training, we estimate $\text{Eff}_{max}(\lambda^{(m)}, \theta^{(m)})$ to be the highest cumulative efficiency calculated from the batches of generated devices.  $\text{Eff}^{(m)}$ is the efficiency of the mth device and is directly calculated with the forward electromagnetic simulation.  The expression $\exp{\left( \frac{\text{Eff}^{(m)} - \text{Eff}_{max}(\lambda^{(m)}, \theta^{(m)})}{\sigma} \right)}$ represents a bias term that preferentially weighs higher efficiency devices during network training and reduces the impact of low efficiency devices that are potentially trapped in undesirable local optima.  The magnitude of this efficiency biasing term can be tuned with the hyperparameter $\sigma$.  A detailed derivation of the loss function from first principles can be found elsewhere \cite{jiang2019dataless} and in the Supporting Information.

The gradient of the loss function with respect to the indices, for the $m$\textsuperscript{th} device, is  $\frac{\partial L}{\partial \mathbf{n}^{(m)}} = -\frac{1}{M}  \exp{\left( \frac{\text{Eff}^{(m)} - \text{Eff}_{max}(\lambda^{(m)}, \theta^{(m)})}{\sigma} \right)} \mathbf{g}^{(m)}$. In this form, minimizing the loss function $L$ is equivalent to maximizing the device efficiencies in each batch.  To train the network and update $\mathbf{w}$, we use backpropagation to calculate $\frac{\partial L}{\partial \mathbf{w}} =  \sum_{m=1}^{M} \frac{\partial L}{\partial \mathbf{n}^{(m)}} \cdot \frac{\partial \mathbf{n}^{(m)}}{\partial \mathbf{w}}$ each iteration.

To ensure that the generated devices are binary, we add $-|\mathbf{n}^{(m)} |\cdot(2-|\mathbf{n}^{(m)}|)$ as a regularization term to the loss function.  This term reaches a minimum when $|\mathbf{n}^{(m)}|=\mathbf{1}$ and the device segments are either silicon or air.  This binarization condition serves as a design constraint that limits metagrating efficiency, as the efficiency enhancement term (Equation 2) favors grayscale patterns.  To balance binarization with efficiency enhancement in the loss function, we include the tunable hyperparameter $\beta$.  Our final expression for the loss function is:
\begin{equation}
    L = -\frac{1}{M} \sum_{m=1}^{M} \Bigg[ \exp{\left( \frac{\text{Eff}^{(m)} - \text{Eff}_{max}(\lambda^{(m)}, \theta^{(m)})}{\sigma} \right)} \mathbf{n}^{(m)} \cdot \mathbf{g}^{(m)} + \beta |\mathbf{n}^{(m)}| \cdot ( 2 -  |\mathbf{n}^{(m)}|) \Bigg]
\end{equation}

We can view conditional GLOnets, in which the non-linear mapping between $(\mathbf{z}, \lambda, \theta)$ and device layout is iteratively improved using physics-driven gradients, as a reframing of the adjoint-based optimization process.  We want to be clear, however, that conditional GLOnets are qualitatively different from adjoint-based topology optimization.  To conceptualize these differences, we discuss each optimization strategy in more detail.  Adjoint-based topology optimization applies to a single device and is a local optimizer.  The algorithm takes an initial dielectric distribution and enhances its efficiency by adjusting its refractive indices at each segment using gradient ascent (Figure 2A).  This method is performed iteratively until the device reaches a local maximum in the design space.  The performance of the final device strongly depends on the choice of initial dielectric distribution \cite{yang2017topology}.  More of the design space can be explored with this approach by performing topology optimization on many devices, each with different initial dielectric distributions.  Devices that happen to have initial dielectric distributions in favorable design space regions will become high performing.

Local optimizers are an effective tool to designing a wide range of photonic devices.  However, their usage is accompanied by a number of caveats.  First, they require significant computational resources.  Hundreds of electromagnetic simulations are required to topology optimize a single device, and for multi-functional devices, this number of simulations can scale to very large numbers.  Second, the sampling of the design space is limited to the total number of devices being optimized.  For complex devices described by a very high dimensional design space, the required sampling may be extremely large.  Third, devices are optimized independently of one another, and gradient information from one device does not impact other devices.  Fourth, multiple topology optimizations are required to produce different devices with different operating parameters.

Conditional GLOnets are qualitatively different in that they optimize an entire distribution of device instances, as mediated by the noise vector $\mathbf{z}$.  The starting point of each conditional GLOnet iteration is similar to adjoint optimization and involves the calculation of efficiency gradients for individual devices using the adjoint method.  However, the difference arises when these gradients are backpropagated into the network.  Consider a single device defined by inputs $(\mathbf{z}, \lambda, \theta)$ that produces a gradient $\mathbf{g}$ for network backpropagation: all the weights in the network get updated, thereby modifying the complete mapping of $\{\mathbf{z}\}$ to $\{\mathbf{n}|\lambda, \theta\}$ (Figure 2B).  This points to the presence of crosstalk between all device instances during the network learning process.  Crosstalk can be useful when devices in promising parts of the design space bias the overall distribution of device instances to these regions.  Regulation of the amount of crosstalk between devices, which is important to stabilizing the optimization method, is achieved from the non-linearity intrinsic to the neural network itself.

Another advantage of our conditional GLOnet is that it is effective at globally surveying the design space, enhancing the probability that optimal regions of the design space are sampled and exploited.  Such global surveying is made possible in part because the initial network supports mapping of $\{\mathbf{z}\}$ onto the full device design space, and in part because different $(\mathbf{z}, \lambda, \theta)$ are sampled each iteration, leading to the cumulative sampling of different regions of the design space during training.  Conditional GLOnets also enable the simultaneous optimization of devices designed for operating parameters that span a broad range of values, over a single network training session.  For our metagratings, these parameters are the outgoing angle and wavelength, but they can generally involve any combination of design parameters in the problem including device thickness, refractive index, or light polarization, amongst others.  This co-design leads to a substantial reduction in computation time per device, which results because these devices operate with related physics and strongly benefit from crosstalk from the network training process.

\section{Results and discussion}
To benchmark devices designed from our conditional GLOnet, we first perform adjoint-based topology optimization on metagratings operating across our desired range of wavelengths and angles.  Details pertaining to this calculation can be found elsewhere \cite{ma2018deep, inampudi2018neural}.  These devices operate across a wavelength range between 600 nm and 1300 nm, in increments of 50 nm, and across a deflection angle range between 40 degrees and 80 degrees, in increments of 5 degrees.  For each wavelength and angle pair, we optimize 500 devices, each with random grayscale patterns serving as initial dielectric distributions.  A total of 200 iterations is performed for each optimization, and the deflection efficiencies of the optimized devices are calculated using a rigorous coupled-wave analysis (RCWA) solver \cite{hugonin2005reticolo}. The efficiencies of the best device for each wavelength and angle pair are plotted in Figure 3A.

With our fully trained conditional GLOnet, we generate 500 devices for each wavelength and angle pair by fixing $(\lambda, \theta)$ at the network input and sampling $\mathbf{z}$ 500 times.  Details pertaining to the architecture and training parameters are in the Supplementary Section.  The efficiencies of the best devices for the same wavelengths and deflection angles displayed in Figure 3A are plotted in Figure 3B. These efficiency values indicate that the best devices from the conditional GLOnet compare well with or are better than the best devices from adjoint-based optimization.  Statistically, 75\% of devices from the conditional GLOnet have efficiencies higher than those from adjoint-based optimization, and 92\% of devices from the conditional GLOnet have efficiencies higher than or within 5\% those from adjoint-based optimization.  While our conditional GLOnet performs well for most wavelength and angle values, it does not optimally perform in certain regimes, such as that at short wavelengths and small deflection angles.  We hypothesize that these nonidealities can be improved with further refinement of the network architecture and training process, and this will be the topic of future study.

The efficiency histograms from adjoint-based topology optimization and the conditional GLOnet, for select wavelength and angle pairs, are displayed in Figure 3C.  A more complete set of histograms is in Figure S2.  The histograms show that adjoint-based topology optimization generates devices with highly variable efficiencies.  This indicates that the initial dielectric distributions of these devices broadly span the design space, and with each device being locally optimized, the result is a set of devices with a wide range of layouts and efficiencies.  The conditional GLOnet-generated devices, on the other hand, tend to have more devices clustered at the high efficiency end of the distribution.  An examination of the layouts of these devices indicate that many have very similar geometries (Figure S3).  This trend is consistent with the objective of the conditional GLOnet, which is to optimize the efficiency of $\{\mathbf{n}|\lambda, \theta\}$.

To help visualize the device optimization process with our conditional GLOnet, we show how the distribution of devices in the design space, together with its corresponding efficiency histogram, evolves over the course of network training.  The devices in this example all operate at $\lambda$ = 900 nm and $\theta$ = 60 degrees, and 100 devices are randomly generated during each iteration of training for visualization.  The high dimensional design space is visualized by performing a principle components analysis on the 500 binary metagratings with $\lambda$ = 900 nm and $\theta$ = 60 degrees produced by adjoint-based optimization (used for Figure 3A) and then reducing the dimensionality of the space to two dimensions.  The results are displayed in Figure 4 and Movie S1.  Initially, the distribution of generated devices is spread broadly across the design space and the efficiency histogram spans a wide range of values, with most devices exhibiting low to modest efficiencies.  As network training progresses, the distribution of generated devices more tightly clusters and the efficiency histogram narrows at high efficiency values.  By the 1000 iteration mark, the generated devices have very high efficiencies and the histogram is strongly skewed towards high efficiency values.  A similar analysis for devices operating with other wavelength and angle combinations is presented in Figure S4.

An examination of total computation time indicates that our conditional GLOnet is computationally efficient at simultaneously optimizing a broad range of devices operating at different wavelengths and angles.  A detailed analysis, presented in the Supporting Information, indicates that the conditional GLOnet uses 10x less computational cost compared to our benchmark adjoint-based topology optimization calculations.  We note that as the number of accessible computing nodes scales up, the efficacy and power of conditional GLOnets can be enhanced by implementing  more simulations in parallel and scaling the batch sizes up.  Such scaling is particularly amenable to existing cloud and server computing infrastructure, which generally enable access to large numbers of computing nodes.

Finally, we show that the generated devices from the conditional GLOnet can be further refined using adjoint-based boundary optimization (Figure 5A).  In this algorithm, $\mathbf{g}$ is calculated by conducting a forward and adjoint simulation, which is consistent with topology optimization. However, we only consider the gradients at the silicon-air boundaries of the device and fix the device reractive indices to be binary throughout the optimization.  For this analysis, we perform 10 iterations of boundary optimization on the highest efficiency generated device for each wavelength and angle pair (Figure 3B).  The final device efficiencies after boundary optimization are shown in Figure 5B and the differential changes in efficiency are shown in Figure 5C.  Most of the efficiency changes are relatively modest and only 4\% of devices have efficiency gains larger than 5\%, indicating that devices from the conditional GLOnet are already at or near local optima.

\section{Conclusions}
In summary, we have shown that conditional GLOnets are an effective and computationally-efficient global topology optimizer for metagratings.  A global search through the design space is possible because the generative neural network optimizes the efficiencies of device distributions that initially span the full design space.  The best devices generated by the conditional GLOnet compare well with the best devices generated by adjoint-based topology optimization.  By conditioning GLOnets with a continuum of operating parameters, ensembles of devices can be simultaneously optimized, further reducing overall computational cost.  Future work will focus on extending conditional GLOnets to other metasurface systems, including aperiodic broadband devices.  The loss function for those design problems can be defined in the same way here, but with $\mathbf{g}$ tailored to the specific optimization target.  For broadband devices, for example, $\mathbf{g}$ should consist of the weighted summation of efficiency gradients at different wavelengths. Given the generality of our approach, we envision that conditional GLOnets can apply to the design of other classes of photonic devices and more broadly to other physical systems in which device performance can be improved using gradients.

\begin{figure}[h]
  \includegraphics[width=\linewidth]{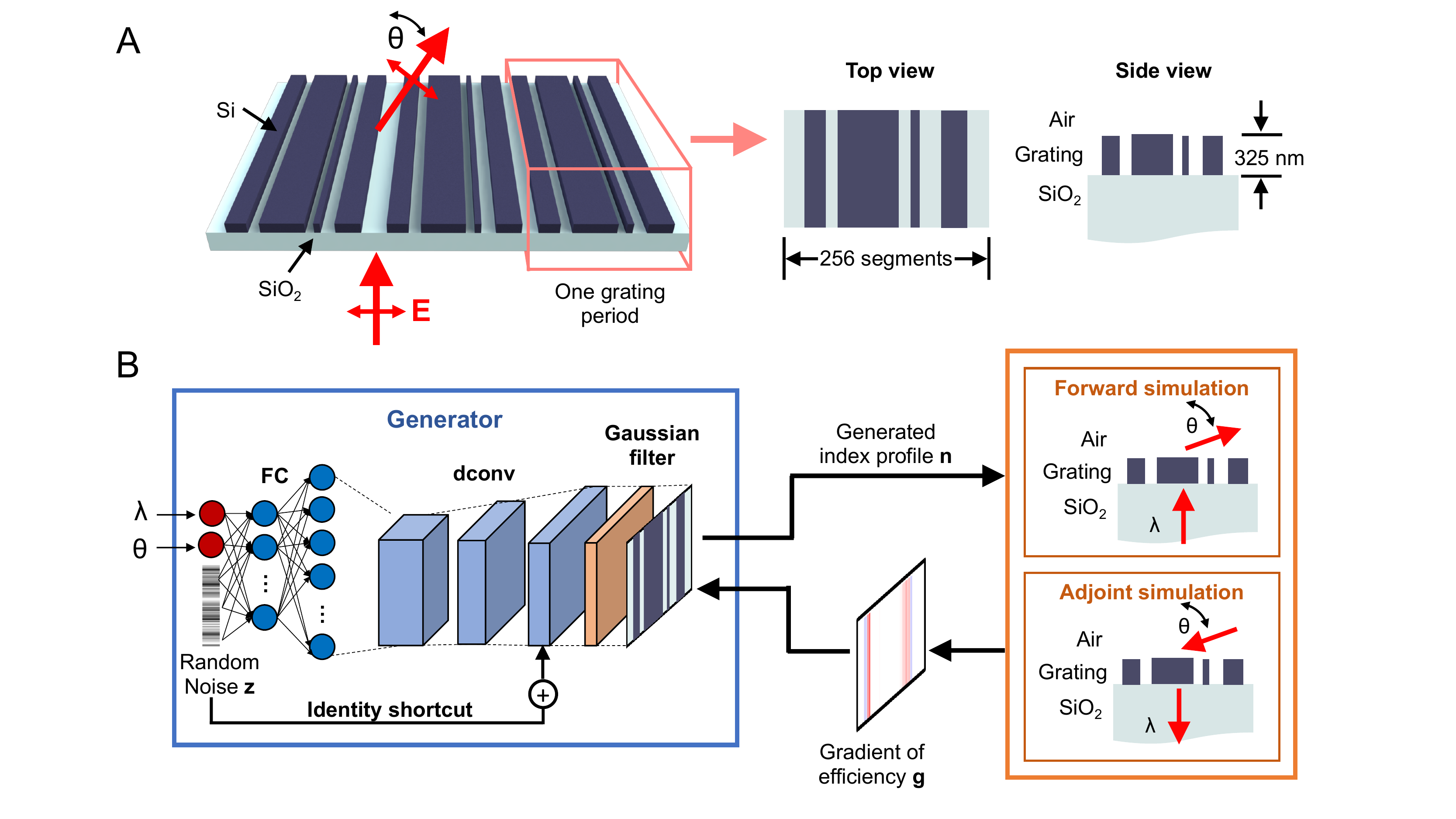}
  \caption{\textbf{Global optimization based on a generative neural network.} \textbf{(A)} Schematic of a silicon metagrating that deflects normally incident TM-polarized light to the outgoing angle $\theta$. The metagrating consists of 325nm-thick Si ridges in air on a SiO\textsubscript{2} substrate.  In the generative neural network, the device is specified by a 1 $\times$ 256 vector, $\mathbf{n}$, which represents the refractive index profile of one period of the grating.  \textbf{(B)} Schematic of the conditional GLOnet for metagrating generation.  The generator is built on fully connected layers (FC), deconvolution layers (dconv), and a Gaussian filter.  An identity shortcut connection is also used and adds z to the output of last deconvolution layer.  The input is the device wavelength $\lambda$, deflection angle $\theta$, and an 256-dimensional noise vector $\mathbf{z}$, and the output is the device vector $\mathbf{n}$. During each iteration of training, a batch of devices is generated and efficiency gradients $\mathbf{g}$ are calculated for each device using forward and adjoint electromagnetic simulations. These gradients are backpropagated through the network to update the weights of the neurons.}
  \label{fig1}
\end{figure}

\begin{figure}[h]
  \includegraphics[width=\linewidth]{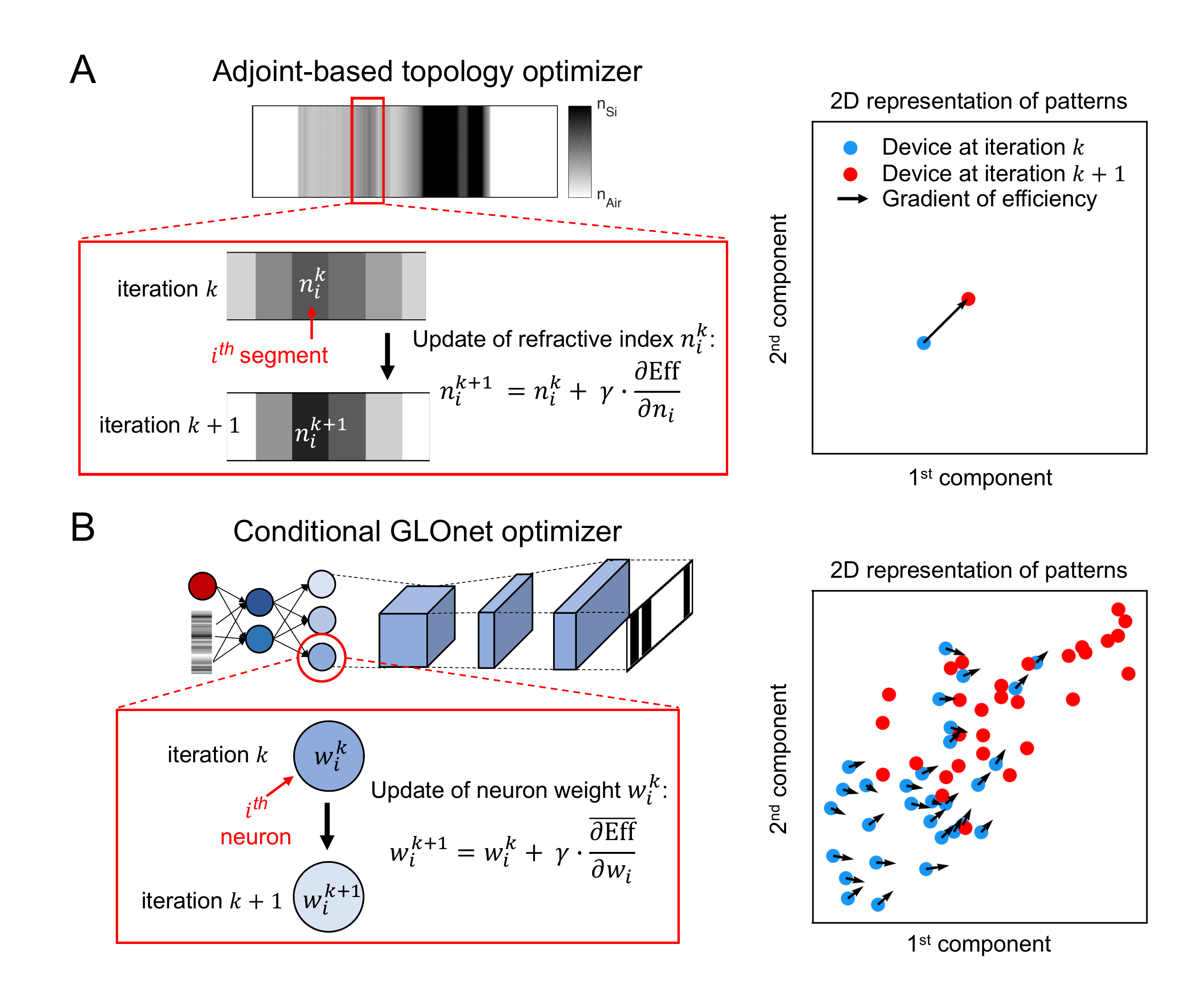}
  \caption{\textbf{Comparison between adjoint-based topology optimization and conditional GLOnet optimization.}  \textbf{(A)} Adjoint-based topology optimization uses efficiency gradients from an individual device to improve its performance within the local design space.  A visualization of the device in a 2D representation of the design space illustrates that from iteration $k$ to $k+1$, the device moves incrementally to a nearby local maxima, indicated by its local gradient.  \textbf{(B)} Conditional GLOnets use a neural network to map random noise to a distribution of devices.  Gradients of efficiency, averaged over a batch of devices, are backpropagated to update the weights of the neurons and deconvolution kernels, which improves the average efficiency of the generated device distribution.  A visualization of the device distribution illustrates that from iteration $k$ to $k+1$, the efficiency gradients from individual devices (black arrows) are used to collectively bias the device distribution towards high efficiency regions of the design space.}
  \label{fig2}
\end{figure}

\begin{figure}[h]
  \includegraphics[width=\linewidth]{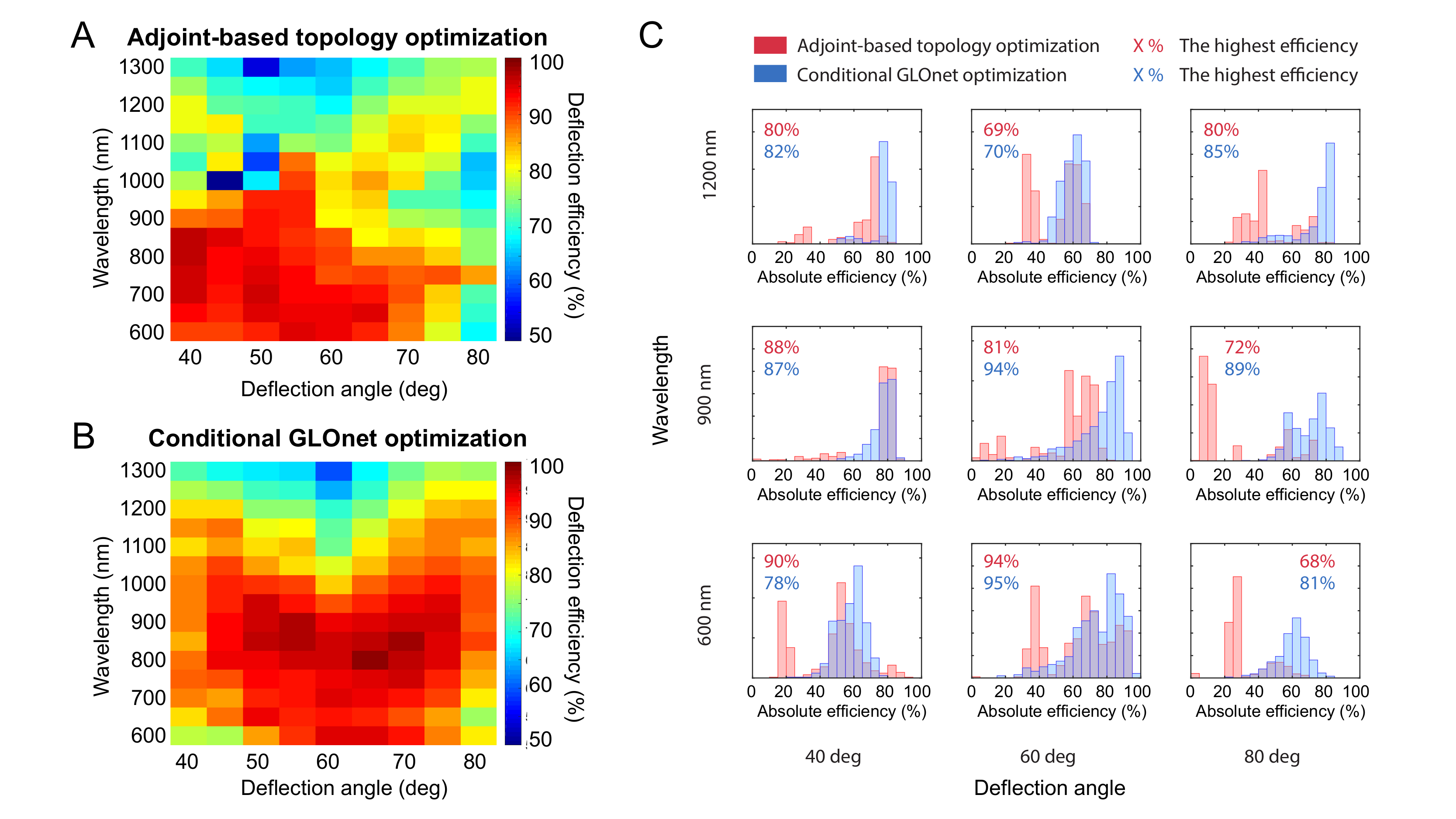}
  \caption{\textbf{Performance comparison of adjoint-based topology optimization and conditional GLOnet optimization.} \textbf{(A)} Plot of metagrating efficiency for devices operating with different wavelength and angle values, designed using adjoint-based topology optimization.  For each wavelength and angle combination, 500 individual optimizations are performed and the highest efficiency device is used for the plot.  \textbf{(B)} Plot of metagrating efficiency for devices designed using the conditional GLOnet.  For each wavelength and angle combination, 500 devices are generated and the highest efficiency device is used for the plot.  \textbf{(C)} Efficiency histograms of devices designed using adjoint-based topology optimization (red) and conditional GLOnet optimization (blue).  The highest device efficiencies in each histogram are also displayed.  For most wavelength and angle values, the efficiency distributions from the conditional GLOnet are narrower and have higher maximum values compared to those from adjoint-based topology optimization. }
  \label{fig3}
\end{figure}

\begin{figure}[h]
  \includegraphics[width=\linewidth]{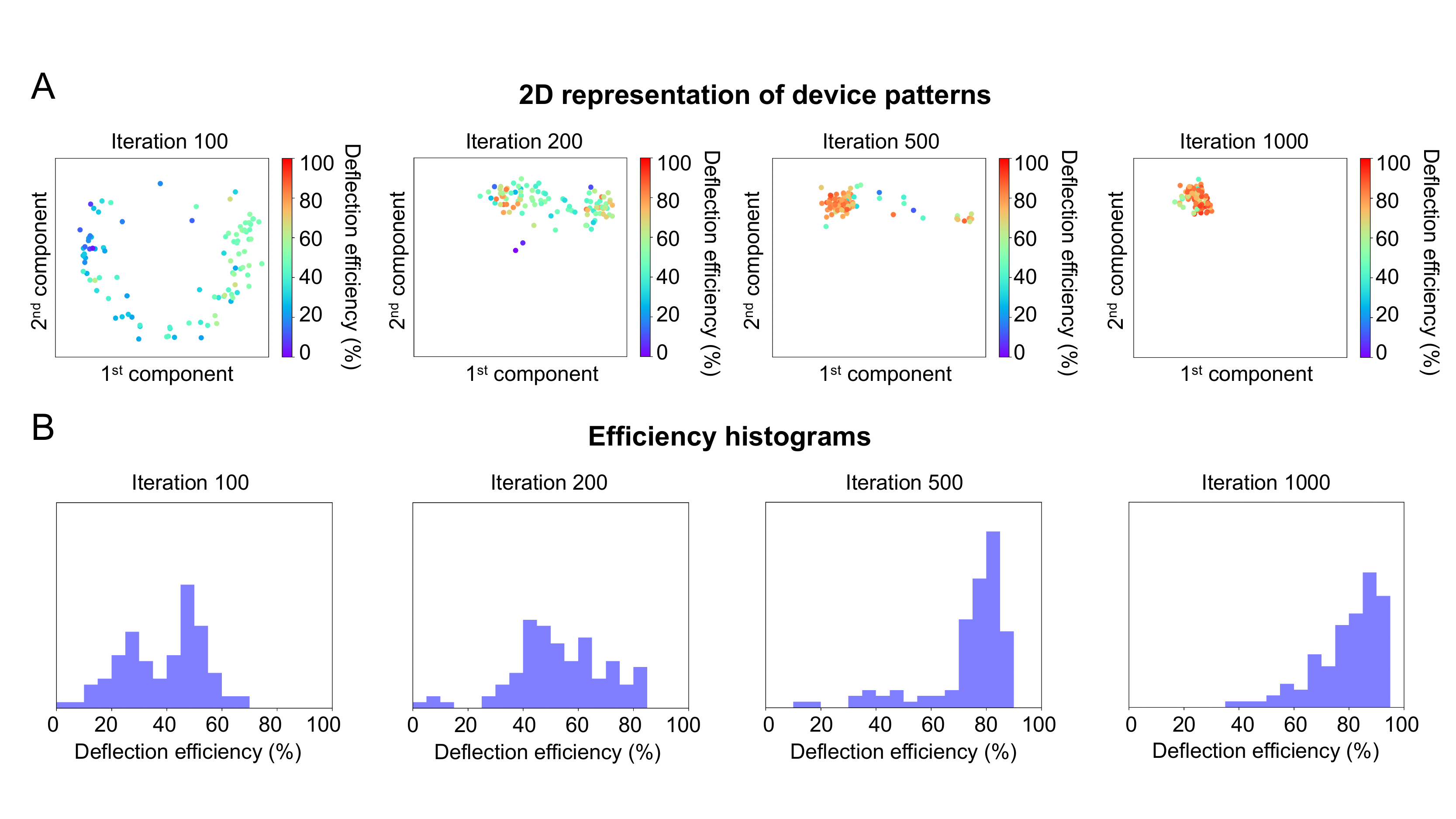}
  \caption{\textbf{Evolution of device patterns and efficiency histograms as a function of conditional GLOnet training.} \textbf{(A)} Visualization of 100 device patterns generated by the conditional GLOnet at different iteration numbers, depicted in a 2D representation of the design space.  All devices are designed to operate at a wavelength of 900 nm and an angle of 60 degrees.  The distribution of generated devices is initially spread out in the design space at the early stages of training and converges to a high efficiency cluster by the 1000 iteration mark.  \textbf{(B)} Efficiency histogram of generated devices at different iteration numbers. The efficiency histogram is initially broad and converges to a distribution of devices biased towards high efficiencies by the 1000 iteration mark.  }
  \label{fig4}
\end{figure}

\begin{figure}[h]
  \includegraphics[width=\linewidth]{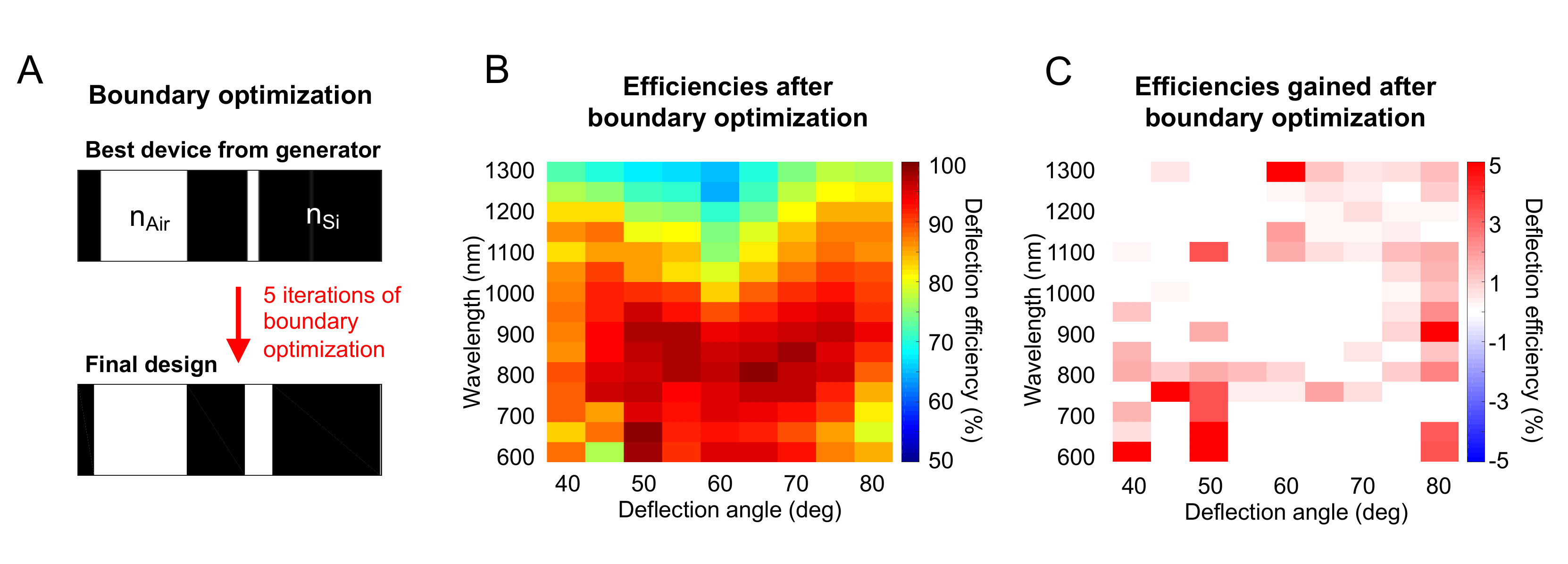}
  \caption{\textbf{Device refinement using boundary optimization.} \textbf{(A)} Boundary optimization uses efficiency gradients to refine the boundaries of binary structures.  \textbf{(B)} Efficiency plot of devices generated by the conditional GLOnet and then refined with 10 iterations of boundary optimization.  For this plot, the best device from Figure 3B for each wavelength and angle combination is used for boundary optimization.  \textbf{(C)} Plot of gains in efficiency after boundary optimization, calculated from the data in Figure 3B and Figure 5B.  Most devices experience modest boosts in efficiency, and 4\% of devices exhibit over a 5\% efficiency improvement. }
  \label{fig5}
\end{figure}

\begin{acknowledgement}

The simulations were performed in the Sherlock computing cluster at Stanford University. This work was supported by the U.S. Air Force under Award Number FA9550-18-1-0070, the Office of Naval Research under Award Number N00014-16-1-2630, and the David and Lucile Packard Foundation. Competing interests: Authors declare no competing interests.

\end{acknowledgement}

\begin{suppinfo}
\begin{itemize}
  \item Network architecture and training process of GLOnet
  \item Formulation of the loss function for the conditional GLOnet
  \item Computational cost of the conditional GLOnet versus adjoint-based topology optimization
  \item Figures S1 to S4
  \item Movie S1
\end{itemize}
This material is available free of charge \textit{via} the Internet at http://pubs.acs.org.
\end{suppinfo}

\bibliography{ref}

\end{document}